\documentclass[10pt,twocolumn,letterpaper]{article}

\usepackage{cvpr}
\usepackage{times}
\usepackage{epsfig}
\usepackage{graphicx}
\usepackage{amsmath}
\usepackage{amssymb}

\usepackage[pagebackref=true,breaklinks=true,letterpaper=true,colorlinks,bookmarks=false]{hyperref}

\cvprfinalcopy 

\def\cvprPaperID{****} 

\ifcvprfinal\fi
\begin{document}

\title{The 1st Challenge on Remote Physiological Signal Sensing (RePSS)}

\author{Xiaobai Li\textsuperscript{1}, Hu Han\textsuperscript{2}, Hao Lu\textsuperscript{2}, Xuesong Niu\textsuperscript{2}, Zitong Yu\textsuperscript{1},  \\ Antitza Dantcheva\textsuperscript{3}, Guoying Zhao\textsuperscript{1}, Shiguang Shan\textsuperscript{2}\\
\textsuperscript{1}Center for Machine Vision and Signal Analysis, University of Oulu, Finland \\  \textsuperscript{2}Institute of Computing Technology, CAS. China \qquad 
\textsuperscript{3}STARS team, INRIA, France\\
\tt\small \{xiaobai.li, zitong.yu, guoying.zhao\}@oulu.fi, \\
\tt\small \ hao.lu@miracle.ict.ac.cn, xuesong.niu@vipl.ict.ac.cn \\
\tt\small\{huhan, sgshan\}@ict.ac.cn, antitza.dantcheva@inria.fr
}

\maketitle

\begin{abstract}
   Remote measurement of physiological signals from videos is an emerging topic. The topic draws great interests, but the lack of publicly available benchmark databases and a fair validation platform are hindering its further development. For this concern, we organize the first challenge on Remote Physiological Signal Sensing (RePSS), in which two databases of VIPL and OBF are provided as the benchmark for kin researchers to evaluate their approaches. The 1st challenge of RePSS focuses on measuring the average heart rate from facial videos, which is the basic problem of remote physiological measurement. This paper presents an overview of the challenge, including data, protocol, analysis of results and discussion. The top ranked solutions are highlighted to provide insights for researchers, and future directions are outlined for this topic and this challenge.
\end{abstract}

\section{Introduction}

Physiological signals such as the heart rate (HR), respiration rate (RR), and heart rate variability (HRV) are important indicators of human physical conditions. Until today most main-stream approaches for measuring physiological signals still reply on contact sensors, including special medical instruments like the electrocardiography (ECG), and some commercial products like sport watches or smart bracelets. To pursue convenient and comfort way for physiological signal measurement, efforts have been made during the last decade for remote measurement from facial videos recorded with commonly accessible cameras. Compared with contact measures, the advantages that remote measures could bring are that, firstly, breaking the constrain of physical distance that people can be measured at different locations; secondly, allowing more comfortable monitoring, especially for patients with special conditions that might be irritated by contact means; and thirdly, integrating the measurement with camera systems which are available almost everywhere in the world. If feasible, remote physiological signal measures would facilitate applications in many fields, e.g., pushing the Telemedicine to another level, which stands high value in the background of global COVID-19 outbreak while the paper was written.

Many methods for remote physiological signal measurement have been proposed ever since 2008, when Verkruysse et al.~\cite{verkruysse2008remote} first reported that plethysmography (PPG) signals can be captured from human faces under ambient light. Of all the physiological signals, the HR was the main focus of most studies, while other signals (e.g., the RR and HRV) were explored in a small number of studies. From the ‘feature’ point of view, remote HR measurement methods can be divided as ‘color-based approaches’ and ‘motion-based approaches’. Color-based approaches such as~\cite{poh2010non},~\cite{poh2011advancements},~\cite{de2013robust},~\cite{li2014remote},~\cite{mcduff2014improvements} and~\cite{wang2017algorithmic} rely on the subtle color changes of facial skin pixels to measure HRs, while motion-based approaches such as~\cite{balakrishnan2013detecting},~\cite{andreia2016TBE} and~\cite{yang2017estimating} track the motion trajectories of facial pixels to measure HRs. From the ‘learning’ point of view, remote HR measurement methods can be divided as ‘training-free approaches’ and ‘learning-based approaches’. Most earlier approaches are training-free, including ~\cite{poh2010non,poh2011advancements},~\cite{de2013robust},~\cite{balakrishnan2013detecting} and~\cite{mcduff2014improvements}, which don’t involve any training process, and rely on signal filtering methods such as blind source separation and others to refine the HR signals. Later studies started to exploit the strength of machine learning or deep learning to further tackle the problem, such studies include~\cite{hamed2014a},~\cite{niu2019rhythmnet},~\cite{niu2018Synrhythm},~\cite{chen2016realsense},~\cite{yu2019remote} and~\cite{yu2019ICCV}. More details of the development of remote HR measurement are referred to survey papers~\cite{philip2018remote} and~\cite{sebastian2018cardiovascular}.

Despite the thriving research interests, the lack of publicly available benchmark databases and a fair validation platform are the major issues that hinder its further development. Kin researchers have to make repetitive efforts on self-collecting small datasets to test proposed methods, which makes it difficult to fairly evaluate and compare the actual strength and weakness of each proposed method, as self-collected data are of different recording conditions and qualities. For this concern, we organize the first challenge on Remote Physiological Signal Sensing (RePSS) in conjunction with the CVPM workshop~\footnote{\url{http://www.es.ele.tue.nl/cvpm20/}} in CVPR 2020 at Seattle, USA. As the first open challenge on remote physiological signal sensing, we will be focusing on measuring the average HR from color facial videos, which is the most fundamental problem in this field.

The rest part of the paper are organized as follows: Section 2 gives the overview of the RePSS challenge, including the data, challenge protocol and evaluation metrics; Section 3 briefly introduces some proposed approaches that achieved leading performance in the challenge, Section 4 reports challenge results and discussions, and at last in Section 5 we discuss future directions.

\section{Challenge Overview}
\subsection{Data}
The data used for the RePSS challenge come from two databases: the VIPL-HR-V2 and the OBF. 

VIPL-HR-V2 is provided by the Institute of Computing Technology (ICT), Chinese Academy of Sciences (CAS), China. VIPL-HR-V2 is the second version of VIPL-HR~\cite{niu2018VIPL-HR,niu2019rhythmnet}, and the construction of VIPL-HR-V2 started from 2018. One important motivation for building VIPL-HR-V2 is to provide large scale data which could meet the need of deep learning methods for the purpose of remote Physiological signal sensing. So far the data of more than 3000 persons were collected in VIPL-HR-V2. The statistical information of the all subjects is listed in Table~\ref{table:subjects}. VIPL-HR-V2 contains facial videos recorded with color cameras under relatively natural ambient light. The subjects were in sitting position in front of recording cameras (Realsense F200) on a table for capturing videos. Ground truth physiological signals of HR, SpO2 and blood volumn pulse (BVP) signals are synchronously recorded with facial videos using a CONTEC CMS60C BVP sensor. Besides, subjects were asked to look as natural as possible during video recording, i.e., allowing head movement and talking.  

OBF is provided by the Center for Machine Vision and Signal Analysis (CMVS), University of Oulu, Finland. The data was collected from 2017 to 2018, which contains data from more than 200 subjects. OBF has subjects of both healthy person and patients with atrial fibrillation (AF), as a main motivation for OBF is to promote remote sensing in medical oriented applications. OBF subjects are from various ethnics including typical eastern Asians (Chinese, Japanese, etc.), Caucasians (Finnish, Russian, Spanish, etc), and others (Indian, Pakistanis, etc.), which means the OBF data covers wide range of skin tones. More statistical information of the 100 healthy subjects are listed in Table~\ref{table:subjects}. Facial videos were recorded with one RGB camera (Blackmagic URFA mini) at 60 fps with resolution of 1920 by 1080, and one NIR camera at 30 fps with resolution of 640 by 480. Three channels of physiological signals (ECG at 256Hz, BVP at 128 Hz, and respiration at 32Hz) were synchronized and recorded with a NeXus-10 MKII platform. Subjects were recorded firstly at resting state and then after five-minutes of intense exercise in order to cover a wider range of HR variance. Heart rate value corresponding to each video is provided, which is the average of all heart rates in the corresponding time period of the video. 

\begin{table}[h]
\caption{Statistical information of VIPL-HR-V2 and OBF subjects.}
\label{table:subjects}
\begin{center}
\begin{tabular}{|c|c|c|}
\hline
 &VIPL-HR-V2&OBF\\
\hline
Age (y)&$35.4\pm 18.0, $&$31.6\pm 8.8, $  \\
 & [6, 60] & [18, 68] \\
\hline
Gender&49\% M , 51\% F&61\% M , 39\% F\\
\hline
Ethnic&Asian:100\%&Caucasian:32\%\\
 & & Asian:37\%,\\
 & & Others:31\%\\
\hline
Weight (Kg)&	$61\pm 12$&	$71\pm 16$\\
\hline
Wear eyeglasses& N/A &39\%\\
\hline
\end{tabular}
\end{center}
\end{table}

\begin{figure*}[htb]                                 
\centering                                      
\includegraphics[scale=0.35]{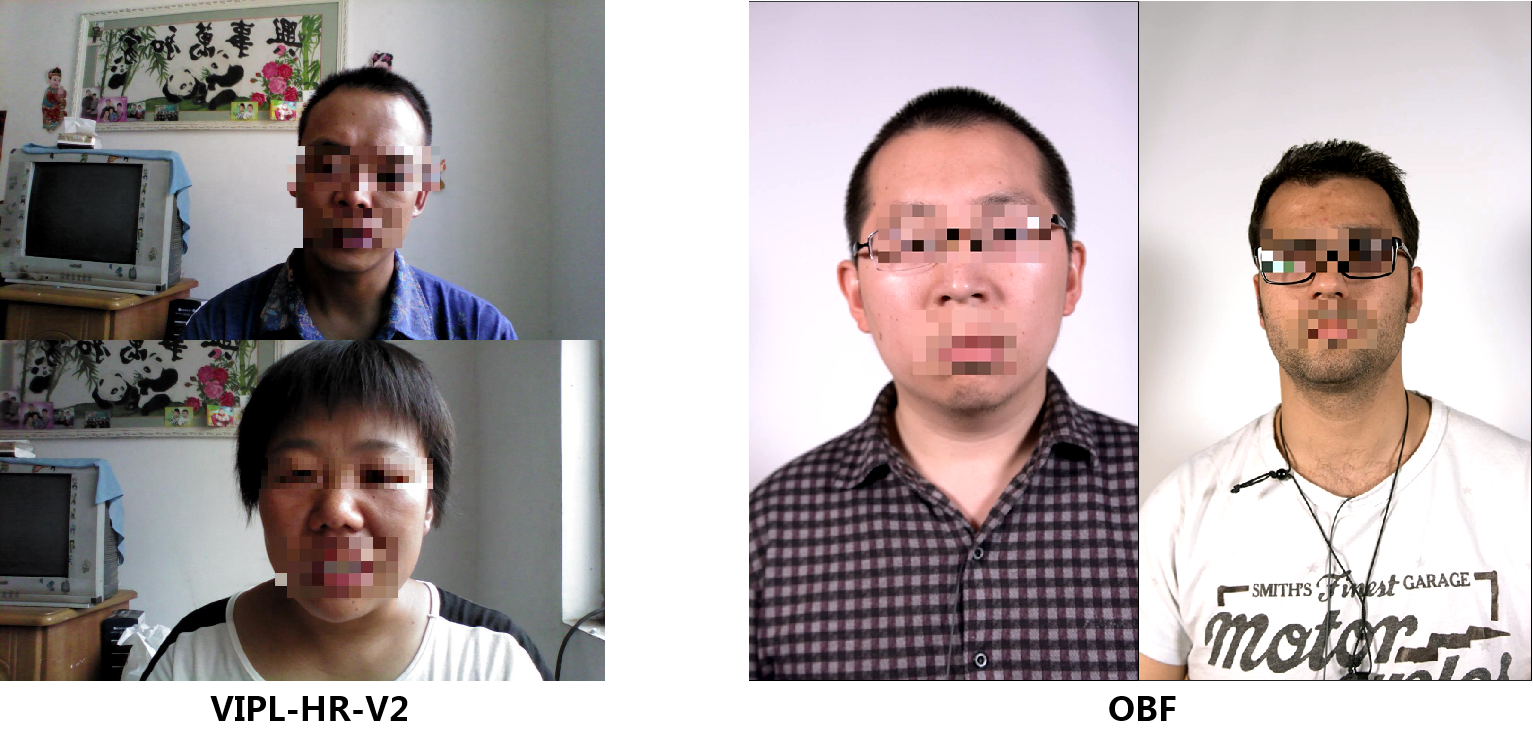}            
\caption{ Sample images of anonymized testing videos. The left one from VIPL-HR, and the right one from OBF.}
\label{fig2} 
\end{figure*} 

\textbf{RePSS challenge training data}
The training data of RePSS are randomly selected from VIPL-HR-V2 database. RGB videos of 500 subjects recorded with Realsense F200 camera at the average speed of 25 fps with resolution of 960 by 720 are used. For each subject we randomly cut five clips of ten-second long videos, so that the training set contains 2500 samples. The ground truth of HR (in beat-per-minute ‘bpm’) is the average of HRs of corresponding clip and provided to challenge participants for their training of models.  

\textbf{RePSS challenge testing data}
The testing data of RePSS challenge consists of two parts, that 100 subjects (no overlap with the training set) from the VIPL-HR-V2 database and 100 subjects (all from the healthy group) from the OBF databases are used. For each subject from both databases, we randomly cut five clips of ten-second long videos, so that the testing set contains 1000 samples. For the VIPL-HR-V2 part, all videos were recorded with the color camera at the average speed of 25 fps with resolution of 960 by 720, and for the OBF part, all videos were recorded with the RGB camera at frame rate of 30 fps (down sampled from the original 60fps to match with the VIPL-HR-V2 data) with resolution of 1920 by 1080. Even though all the participating subjects have signed consent forms and given the permission to use all the recorded data for scientific research and demonstrations in e.g., publications and presentations, the OBF videos were anonymized by adding mosaic blocks covering important facial features to better protect the personal identification while data is being used for research. Face positions and facial landmark locations were detected using face-alignment~\footnote{\url{https://github.com/1adrianb/face-alignment}} and were provided for challenge participants to facilitate the testing process if needed. The testing data from VIPL-HR-V2 were processed in the same way to unify the format. Sample images of the anonymized testing videos are shown in Figure~\ref{fig2}. Ground truth average HRs were computed from corresponding BVP signals of both databases, which were not provided to challenge participants and only be used for the evaluation carried out the challenge organizers based on the results submitted from the participants.

\subsection{Challenge protocol}

The RePSS challenge is operated on the CodaLab platform, and consists of two stages as follows.

\textbf{Training phase (15.01.2020 – 20.02.2020)}
The training data was released on 15th. Jan. 2020. During the training phase, registered participants get access to the labelled training data and establish their machine learning models. There was no specific limitation of using outer source data, i.e., if some participants want, they can also use their own data. No result submission could be made to the challenge website during the training phase. 

\textbf{Testing phase (23.02.2020 – 06.03.2020)}
The testing data was released on 23rd. Feb. 2020. During the testing phase, challenge participants were asked to adjust their models using the testing data and submit testing results to the challenge website to check the performance. Test results were asked to be submitted in the form of an excel table which contains estimated average HR for each test sample. The ground truth HRs were embedded in the CodaLab platform to automatically produce final performance when a new result was submitted. Executable codes were not asked for this challenge, but may be considered in future. Each registered participant (or team) can submit results up to five times before the submission deadline, and the best performance (of the participant or the team) will be ranked and shown in the final result leading board. It is possible that one participant could register under multiple names though.

\subsection{Evaluation /metrics}
Three evaluation metrics were used for the RePSS challenge, including the mean average error (MAE), the root mean square error (RMSE) and the Pearson’s correlation efficient r (R). All three metrics are widely used in related papers. The MAE and RMSE can evaluate the approaches by showing the ‘difference’ of estimated HRs compared to the actual HRs on an average level, thus smaller value indicates better performance; while the correlation R shows how strongly the relationship is on scale of [-1 1], of the estimated HR and the corresponding GT HRs, thus larger R indicates better performance. 

\section{Proposed approaches}
Altogether 129 teams (registered CodaLab names) from 36 organizations all over world participated the first RePSS challenge, and all participants signed license agreements for data access. No constraint was put on which category of method to be preferred or forbidden as long as they can work for remote HR measure. Three approaches from the top three ranked teams are introduced in the following.

\subsection{Mixanik (Neurodata Lab)}

\begin{figure}[h]                                 
\centering                                      
\includegraphics[scale=0.3]{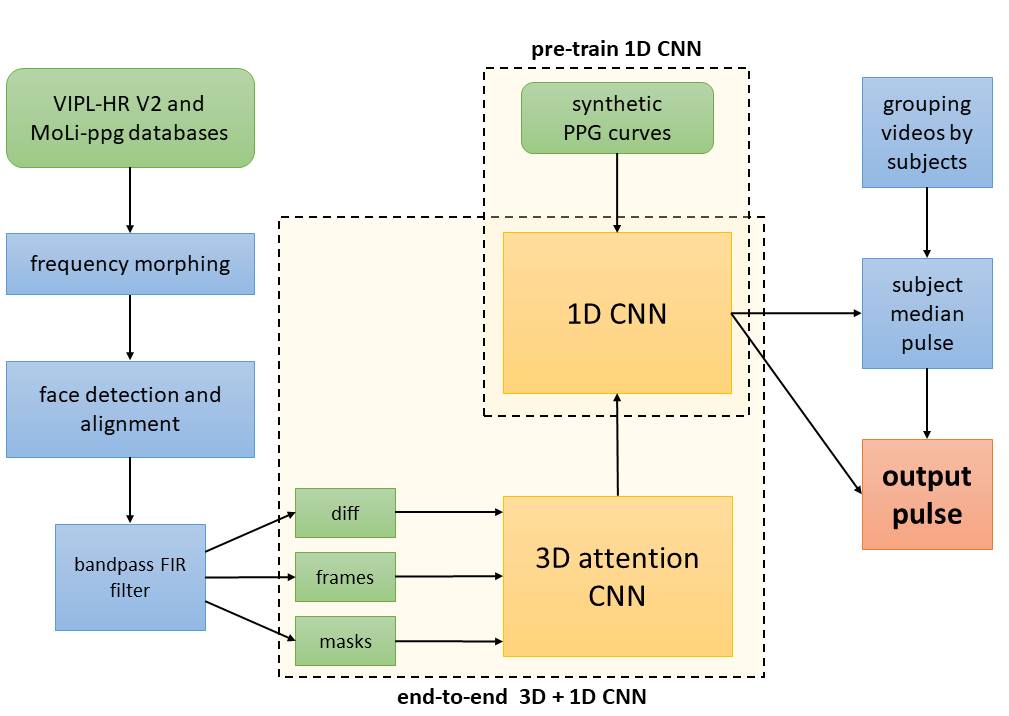}            
\caption{ Overview of Mixanik method.}
\label{fig:rank1} 
\end{figure}

The overview of Mixanik method is shown in Figure~\ref{fig:rank1}.

\textbf{Data augmentation:} speed-up and slow-down augmentation (or frequency morphing)~\cite{niu2019robust} is used to increase training dataset size as well as variance of the reference pulse rate distribution. This should improve the performance of the algorithm especially for subjects with very low or very high pulse rate. Horizontal flip augmentation is used as well.

\textbf{Video preprocessing:} First, the method detects faces using a RetinaNet network~\cite{lin2017focal} with MobileNet backbone~\cite{howard2017Mobilenets} trained with focal loss. Affine face alignment based on facial landmarks detection~\cite{dong2018super} is performed for each face. ROI average pooling is used to resize facial areas to the size of W$\times$H for the heart rate estimation neural network, where $W$=$H$=36. After that, resampling to 25 fps by cubic interpolation is performed. Bandpath filter for [45 bpm, 180 bpm] frequencies is applied for each (pixel, channel) pair independently in order to filter out signals not related to pulse cycles.

\begin{figure*}[t]                                 
\centering                                      
\includegraphics[scale=0.35]{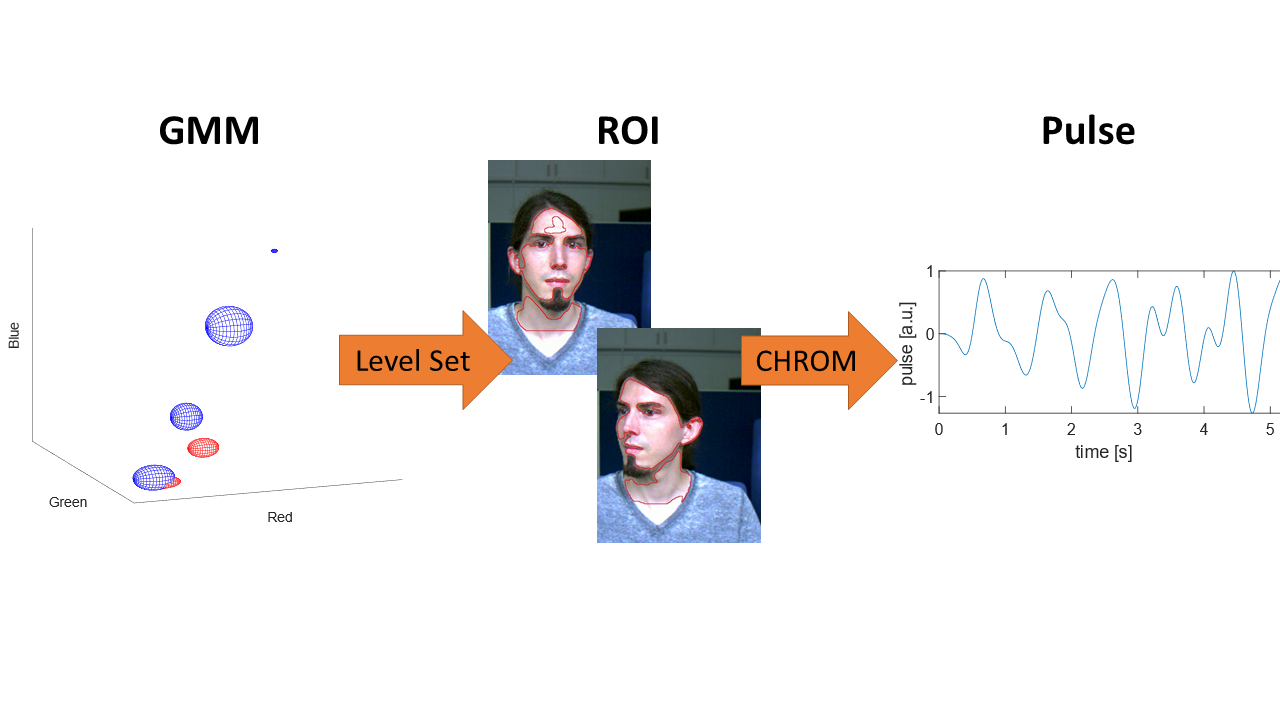}       
\caption{ Overview of AWoyczyk method. Skin GMM (red) and non-skin GMM (blue), segmented ROI after applying level set and extracted pulse wave via CHROM.}
\label{fig:rank2} 
\end{figure*} 

\textbf{Pulse rate estimation neural network:} A convolutional neural network is trained for pulse rate estimation. It has 3 inputs: 1). diff input is a discrete time derivative of the preprocessed frames sequence described above. 2). frames input consists of the preprocessed frames themselves. 3). masks consists of frame-wise masks. These masks are based on the facial landmarks. Each mask pixel equals 0 if the corresponding pixel does not belong to face or belongs to mouth or eyes area, and 1 otherwise. A 3D spatio-temporal attention CNN is used followed by global spatial average pooling for PPG features extraction. Diff input is processed with the 3D CNN with separable spatio-temporal 3D convolutions and spatial pooling layers. Frames and masks are used for attention weights evaluating to select most relevant face areas for pulse rate estimation. 3D CNN outputs 32 time series, one for each channel of the last convolutional layer. Each time series is processed with a 1D CNN, which was pre-trained to evaluate pulse rate on synthetic PPG-like curves. 32 pulse rate estimations are achieved, which are combined to a single output with a 2-layer perceptron. The whole 3D+1D CNN was trained end-to-end with MAE loss function on our MoLi-ppg dataset [in press] (~11.5 hours, 40 subjects) and then or the VIPL-HR V2 training data. Adam optimizer is used during training.

\textbf{Predictions post-processing:} There are 5 video fragments featuring each subject in the competition dataset. According to the training set, these 5 fragments had nearly the same reference pulse rate for most of the subjects. Let ($p_1, p_2, p_3, p_4, p_5$) be the neural network outputs on these fragments for some subject. Then the final pulse rate estimation on a video fragment is $f_i=0.01 \times p_i + 0.99 \times median$ ($p_1, p_2, p_3, p_4, p_5$). The fragments are not grouped by subjects in the test set. To evaluate median value each video was matched with other videos of the same subject. For this purpose the researchers first evaluate a simple embedding of the first frame for each video. This embedding for VIPL dataset videos consists of RGB colors of pixels of two $100 \times 150$ rectangles (top-left and top-right), each resized to $10 \times 15$. So, VIPL videos embedding have length $2 \times 10 \times 15 \times 3$ = 900 and represent background color information. All OBF videos have the same background, so for OBF videos chest area (bottom $420 \times 1080$ pixels rectangle resized to $8 \times 20$) is used as a color embedding. The researchers use $1- \rho(a,b)$ as a distance metric on the embeddings described above, where $\rho$ is a Pearson correlation coefficient. Videos are grouped by subjects by an iterative DBSCAN procedure. First the researchers set $\epsilon$ =0.01 in DBSCAN, and then gradually increase it up to 0.4. If there are any clusters of size 5 on each step, it is assumed that each of these clusters corresponds to videos of one subject. These videos are not considered on the subsequent clustering iterations. 

\subsection{AWoyczyk (Fachhochschele Dortmund)}
The overview of AWoyczyk method is shown in Figure~\ref{fig:rank2}.

The examination of vital parameters is an essential element of modern medicine. The heart rate is one of the most important parameters. It is typically recorded via electrocardiography or by photoplethysmography (PPG) using sensors attached to the patient. Current research focuses on non-contact alternatives to capture physiological signals. One promising approach uses videos to derive a pulse signal (imaging PPG, iPPG). A common approach to derive the heart rate by iPPG first defines a region of interest (ROI) and secondly combines the colour information from that ROI to yield a pulse signal. To define the ROI, simple face detectors producing a facial bounding box, more complex variants yielding facial landmarks and skin classifiers, respectively, are in use. However, previous research has shown that homogeneous skin areas contribute to a better signal. Trumpp et al. therefore presented a level set segmentation to identify a homogeneous skin region~\cite{trumpp2017skin}. Since the level set segmentation described by Chan et al.~\cite{chan2001active} does not guarantee to segment skin from non-skin but merely fore- from background, this contribution adopts the method to the special case of skin segmentation. The researchers propose an approach using a Gaussian mixture model (GMM) based level set formulation to yield a time-varying and homogeneous ROI on which further iPPG processing steps can build up.

The researchers model the probability distributions for the pixel skin and no-skin class by two separate GMM. They are trained on the first frame using the expectation maximation algorithm. The skin GMM originates from the ROI of a face detection algorithm (OpenCV’s Viola \& Jones Face detection) while the surrounding pixels define the non-skin GMM. The proportion of the posterior probabilities are than included in the energy term minimized by the level set function, i.e. the inclusion of non-skin pixels according to the individualized model is penalized, as well as skin pixels outside the ROI. In order to keep track of movements and facial expressions, the level set function is updated for each frame. This procedure yields a time-varying ROI on which the further processing builds up. To derive the pulse signal from the ROI, we use CHROM~\cite{de2013robust}. CHROM uses a combination of normalized chrominance signals, derived from the red, green and blue channel to make the signal more robust to intensity changes originating from motion or reflectance. The CHROM signal is further processed by a bandpass filter. Finally, the heart rate is determined as a frequency belonging to the highest amplitude in the frequency spectrum of the extracted pulse signal.

\subsection{PoWeiHuang (National Chiao Tung University)}

\begin{table*}[]
 \caption{The final result leaderboard of the 1st challenge of RePSS.} 
 \label{fig:rank}
 \centering
\begin{tabular}{|l|l|l|l|l|l|}
\hline
\# & User         & Institute & MAE           & RMSE          & R             \\ \hline
1  & Mixanik      & Neurodata Lab         & 6.94289 (1)   & 10.68021 (1)  & 0.75493 (1)   \\ \hline
2  & AWoyczyk     & Fachhochschele Dortmund         & 7.92115 (2)   & 14.37509 (3)  & 0.58891 (2)   \\ \hline
3  & PoWeiHuang   & National Chiao Tung University         & 8.94626 (3)   & 14.16263 (2)  & 0.53531 (3)   \\ \hline
4  & SHLAI        & National Tsing Hua University         & 12.38949 (4)  & 16.08538 (7)  & 0.22547 (5)   \\ \hline
5  & yuchun\_wang & National Tsing Hua University         & 12.46439 (5)  & 16.20117 (10) & 0.18898 (7)   \\ \hline
6  & Simplar      & Southern Federal University         & 12.48682 (6)  & 15.83572 (4)  & 0.18548 (8)   \\ \hline
7  & yangyb       & N/A         & 12.53826 (7)  & 16.08765 (8)  & 0.10077 (14)  \\ \hline
8  & mayanbiao    & N/A         & 12.54743 (8)  & 16.06358 (6)  & 0.10139 (13)  \\ \hline
9  & liyuxin      & N/A         & 12.57556 (9)  & 15.97100 (5)  & 0.09694 (16)  \\ \hline
10 & yaoguorun    & N/A         & 12.73557 (10) & 16.12572 (9)  & 0.09998 (15)  \\ \hline
11 & cvlab.nthu   & N/A         & 12.73808 (11) & 16.34883 (12) & 0.21761 (6)   \\ \hline
12 & shaoguowen   & N/A         & 12.90892 (12) & 16.54261 (13) & 0.08033 (17)  \\ \hline
13 & chenggj      & PingAn Health Technology Co.Ltd         & 12.91462 (13) & 16.57148 (14) & 0.15358 (10)  \\ \hline
14 & legal        & N/A         & 12.91462 (13) & 16.57148 (14) & 0.15358 (10)  \\ \hline
15 & dlavender    & Nanjing University of Science and Technology         & 13.01585 (14) & 17.30358 (17) & 0.06710 (18)  \\ \hline
16 & cpi1976      & CanControls GmbH         & 13.25897 (15) & 16.28846 (11) & 0.00000 (23)  \\ \hline
17 & baoqianyue   & N/A         & 13.28824 (16) & 16.74968 (16) & 0.02195 (21)  \\ \hline
18 & lg920810     & PingAn Health Technology Co.Ltd         & 13.39697 (17) & 17.86813 (19) & 0.27874 (4)   \\ \hline
19 & mengtzu.chiu & National Tsing Hua University         & 13.55358 (18) & 17.45627 (18) & 0.16305 (9)   \\ \hline
20 & lijingjdsun  & N/A         & 13.63316 (19) & 16.65428 (15) & 0.05501 (19)  \\ \hline
21 & CCCCoda      & Beihang Univercity        & 14.36682 (20) & 18.75101 (20) & 0.11018 (12)  \\ \hline
22 & ylin         & N/A         & 14.50666 (21) & 18.97367 (21) & 0.11282 (11)  \\ \hline
23 & WeihuaOu     & Guizhou Normal University         & 14.75637 (22) & 19.10806 (22) & 0.04032 (20)  \\ \hline
24 & sunrise      & Shanghai Jiao Tong University         & 15.68960 (23) & 19.74385 (23) & 0.00932 (22)  \\ \hline
25 & wantsjean    & N/A         & 20.12449 (24) & 25.55460 (24) & -0.02370 (24) \\ \hline
\end{tabular}
\end{table*}

The method was proposed originally for the purpose of remote monitoring of driving scenarios, which was then adapted to the task of remote HR measurement for attending the RePSS challenge. Based on statistical signal processing (SSS) and Monte Carlo simulations, the researchers propose a new algorithm, ADaptive spectral filter banks (AD), which provides better balance to robustness and sensibility of remote monitoring for driving scenarios. HR estimation with rPPG can be approximately modeled as single-tone frequency estimation with additive white noise. This estimation problem has been discussed thoroughly in the SSS and the probability of outliers can be derived from corresponding signal-to-noise ratio (SNR). Based on the probability of outliers, the method provides a viable spectral filter option to balance the robustness and sensibility. In the design of AD filter banks, the exponential smoothers are selected due to the simple relationship between time constant and design of parameters. If the SNR is high enough and the probability of outlier is tolerable, the time constant of AD is small to enhance tracking sensibility; by contrast, if the SNR is low, large time constant is applied for stability. In addition, because the design is based on SSS and Monte Carlo simulation, the method has a potential advantage over applications with different band-width or applications with different requirement between sensibility and stability.

The researchers built a driving database to verify the proposed algorithm and analyzed the influence on rPPG from drivers’ habits (amateur and professional), vehicle types (compact cars and buses), and routes. In total, a driving database with over 23 hours of data and 104 trials has been built. Moreover, the researchers also adapt their method to the RePSS challenge.

\section{Challenge results and discussion}
In this section we report the results obtained by participating teams. First, the main results are reported and shown in the ranked leaderboard. Then we compare results achieved on the two databases of VIPL-HR-V2 and OBF. At last we analyze the performance on different HR ranges. The results from the top six groups are shown for the last two analysis due to limited space.

\subsection{The main results and ranking}
The main results are ranked with the metric of MAE, and we also calculated two other metrics of RMSE and correlation R in order to evaluate the methods on a fuller scope. The ranking leaderboard is shown in Table~\ref{fig:rank}. For each registered name, up to five submissions can be made and the system chooses the submission and ranked the highest among the five. The best results were achieved by ‘Mixanik’ with an MAE of 6.94 bpm. ‘Mixanik’ also leads on the other two metrics, with the RMSE of 10.68 bpm and R of 0.75.

To further evaluate the performance under different conditions, we choose the results from the top three ranked teams (named as T1, T2 … T6 accordingly) to carry out two comparison analysis in the following two subsections. 

\begin{figure}[h]                                 
\centering                                      
\includegraphics[scale=0.45]{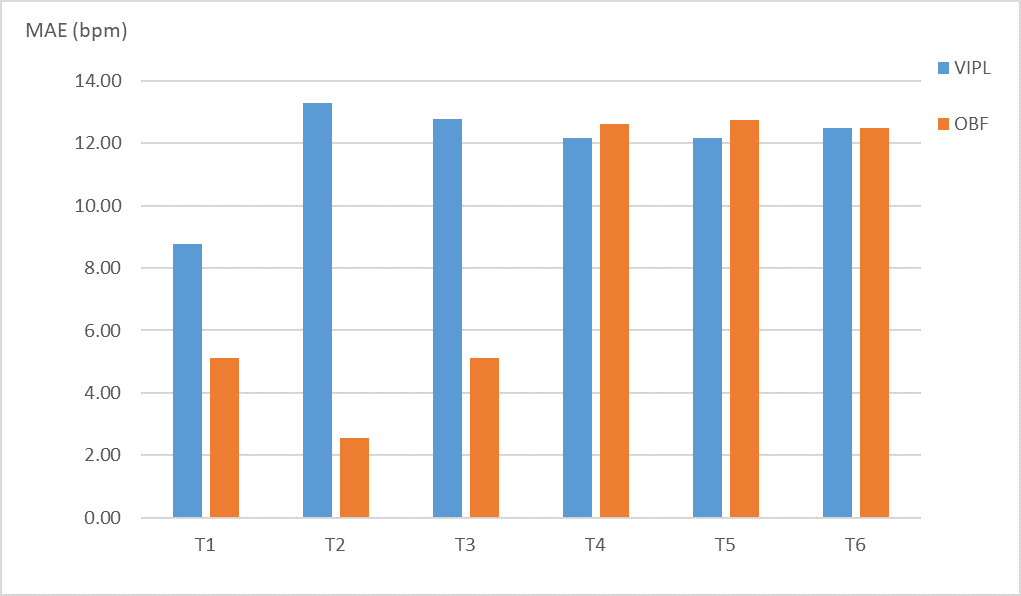}           
\caption{ Comparison of the performance on VIPL-HR-V2 and OBF of the top six teams.}
\label{fig:T16} 
\end{figure} 

\subsection{ Performance on the two databases}
The testing data includes two halves, i.e., 500 samples from the VIPL-HR-V2, and the rest 500 samples from the OBF. We would like to compare the performance on the two parts of data. The metric of MAE was calculated separately on VIPL-HR-V2 and OBF for each team, and the results of T1 to T6 are shown as a bar chart in Figure~\ref{fig:T16}.

The results show that for the top three teams, their methods performed significantly better on the OBF than on the VIPL-HR-V2 data. The best MAE on OBF was 2.56 bpm achieved by T2. The differences are much smaller for T4, T5 and T6. One reason for the difference might be that the OBF videos have higher resolution than the VIPL-HR-V2 videos, which may indicate that the top three approaches are more sensitive to the face size or input resolution.

\subsection{Performance on different HR ranges}
\begin{figure}[h]                                 
\centering                                      
\includegraphics[scale=0.45]{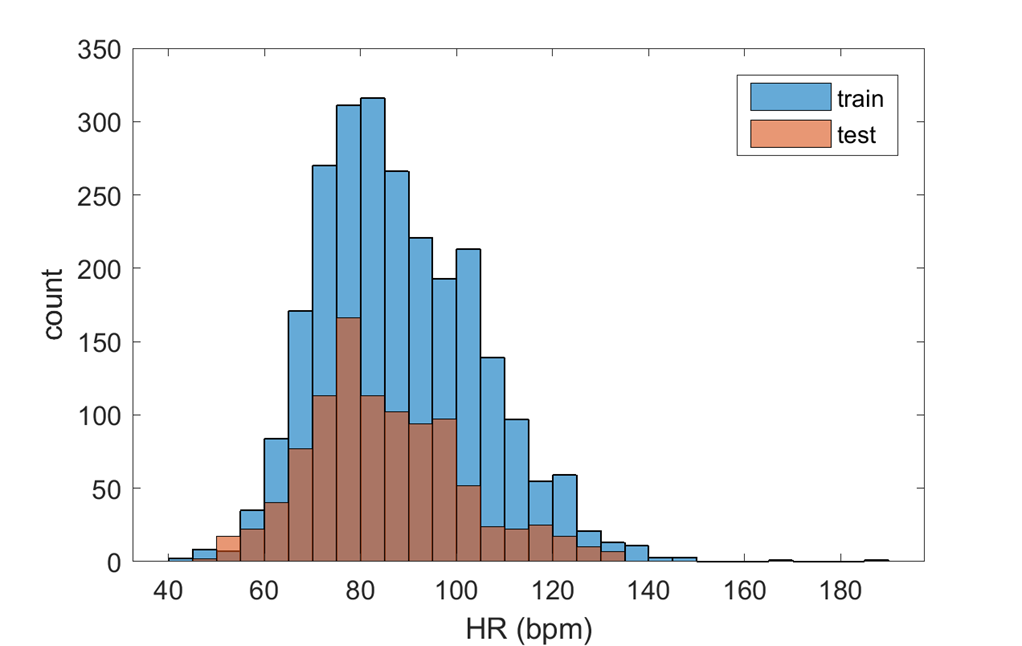} 
\caption{ Distribution of HR levels of the RePSS training and testing data. }
\label{fig:distribution} 
\end{figure}

Healthy adults’ HRs distribute in the range of [50, 130] bpm in most daily life scenarios when middle to low level intensity of activities are involved. The distribution of HRs of our training and testing samples are shown in Figure~\ref{fig:distribution}. The distribution of the testing data match similar pattern of the training data. Our test data covers the range of [49, 134] bmp, which makes a good representation of ordinary HR cases. 

\begin{figure}[htb]                                 
\centering                                      
\includegraphics[scale=0.45]{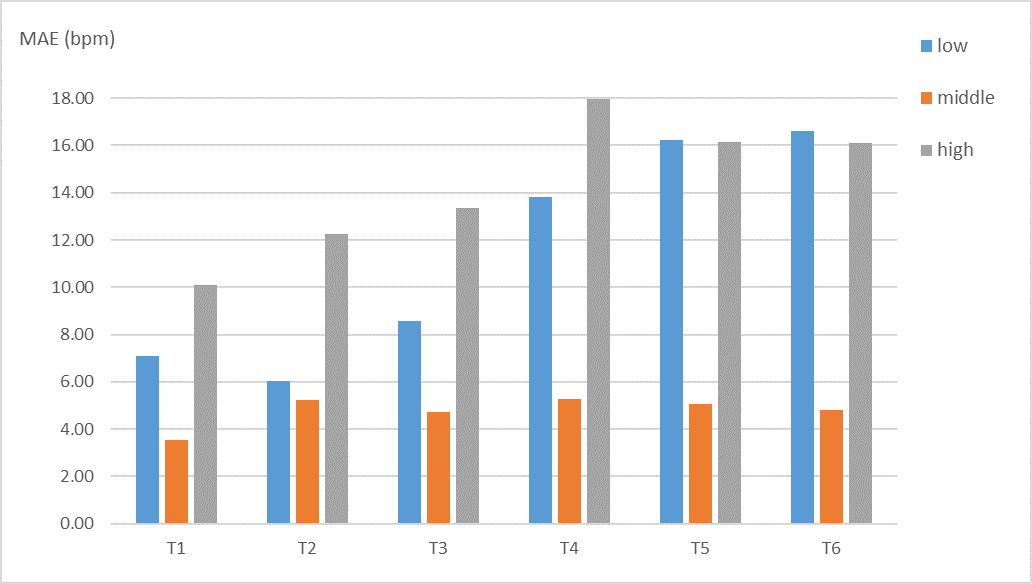}       
\caption{ Comparison of the performance on low vs. middle vs. high HR levels of the top six teams. }
\label{fig:HRrange} 
\end{figure}

We divided the testing samples into three groups (of similar number of samples) of low (less than 77 bpm), middle (77 to 90 bpm) and high (more than 90 bpm) HR levels according to the GT HR values, so that we can examine how well the approaches performed on different HR levels. The MAE values of the three HR groups are calculated for each of the top six teams, and the results are shown in Figure~\ref{fig:HRrange}. It can be seen that all teams performed the best on the middle-level group of data, i.e., ranged in [77, 90] bpm, while the MAE values are significantly larger when tested on either high-level or low-level groups of data. The challenge of measuring high or low level of HRs needs to be addressed in future works.

\section{Future directions}
As the very first challenge held on the topic of remote physiological signal sensing, the RePSS attracted great interests within short time. Registered teams come from various countries and regions (e.g., China, Russia, Germany, USA, Australia, and etc.), and more than one hundred results were submitted at the end, which shows that it is a widely concerned topic. As the first trial, we started with the basic task of measuring average HR from color facial videos. Meanwhile, we provided large amount of training and testing data with non-overlap subjects, and concerned different recording scenarios (e.g., talking, moving, and different lighting). By these means we increase the challenge difficulty level, and make the task more resembling to applications in real world. Very good performance were achieved thanks to the efforts of all participating teams, especially of the top three groups. The MAE values of about 7 bpm is a good starting point, considering that the testing data include masked faces, and are even from different recording sources.

We expect to continue with the RePSS challenge in the following years. We expect that more advanced approaches could be developed to further improve the HR measurement accuracy, i.e., achieve smaller MAEs and higher Rs. Moreover, we will also consider other aspects to make the challenge better, which include:

1). Increase the data size.

2). Enrich data for special concerns, e.g., data with higher or lower HR levels, data from darker skin tones, etc.

3). To have more than one test channels focusing on different tasks, so that teams can join and choose their favorite. 

4). Invite leading teams from institutions or companies in this domain to increase the visibility of RePSS challenge.

In terms of future directions for the RePSS challenge, the analysis of the current results also gave us good hints. First, the measurement of average HR will continue to be a major focus, as the accuracy can be further improved. We may include the measurement of HRV features to elevate the difficulty level. Second, we might set test specially focused on facial resolutions to explore the impact of face size, and hopefully some approaches that could counter for the disadvantage of low resolution would appear. Third, we consider adding the task of measuring respiration rate as it is also an important vital sign in many application scenes. We would also like to hear ideas and concerns from participants, and hope the RePSS challenge could develop and thrive to be a better platform supporting this topic.

\section{Acknowledgement}
The first challenge of RePSS was funded by Academy of Finland and University of Oulu, and Chinese National Natural Science Foundation.

{\small
\bibliographystyle{ieee_fullname}
\bibliography{egbib}

\begin{thebibliography}{10}\itemsep=-1pt

\bibitem{balakrishnan2013detecting}
Guha Balakrishnan, Fredo Durand, and John Guttag.
\newblock Detecting pulse from head motions in video.
\newblock In {\em Proc. IEEE CVPR}, pages 3430--3437, 2013.

\bibitem{chan2001active}
Vese L.~A. Chan, T.~F.
\newblock Active contours without edges.
\newblock {\em IEEE Trans. Image Processing}, 10(2):266--277, 2001.

\bibitem{chen2016realsense}
Jie Chen, Zhuoqing Chang, Qiang Qiu, Xiaobai Li, Guillermo Sapiro, Alex
  Bronstein, and Matti Pietik{\"a}inen.
\newblock {RealSense} = real heart rate: Illumination invariant heart rate
  estimation from videos.
\newblock In {\em Proc. IEEE IPTA}, pages 1--6, 2016.

\bibitem{de2013robust}
Gerard De~Haan and Vincent Jeanne.
\newblock Robust pulse rate from chrominance-based rppg.
\newblock {\em IEEE Trans. Biomed. Eng.}, 60(10):2878--2886, 2013.

\bibitem{dong2018super}
Yu~S. I. Weng X. Wei S. E. Yang Y. Sheikh~Y. Dong, X.
\newblock Supervision-by-registration: An unsupervised approach to improve the
  precision of facial landmark detectors.
\newblock In {\em Proc. IEEE CVPR}, pages 360--368, 2018.

\bibitem{hamed2014a}
Monkaresi Hamed, Rafael~A. Calvo, and Yan Hong.
\newblock A machine learning approach to improve contactless heart rate
  monitoring using a webcam.
\newblock {\em IEEE JOURNAL OF BIOMEDICAL AND HEALTH INFORMATICS}, 18(4), 2014.

\bibitem{howard2017Mobilenets}
Zhu M. Chen B. Kalenichenko D. Wang W. Weyand T. Adam~H. Howard, A.~G.
\newblock Mobilenets: Efficient convolutional neural networks for mobile vision
  applications.
\newblock In {\em arXiv}, 2017.

\bibitem{li2014remote}
Xiaobai Li, Jie Chen, Guoying Zhao, and Matti Pietikainen.
\newblock Remote heart rate measurement from face videos under realistic
  situations.
\newblock In {\em Proc. IEEE CVPR}, pages 4264--4271, 2014.

\bibitem{lin2017focal}
Goyal P. Girshick R. He K. Dollár~P. Lin, T.~Y.
\newblock Focal loss for dense object detection.
\newblock In {\em Proc. IEEE CVPR}, pages 2980--2988, 2017.

\bibitem{mcduff2014improvements}
Daniel McDuff, Sarah Gontarek, and Rosalind~W Picard.
\newblock Improvements in remote cardiopulmonary measurement using a five band
  digital camera.
\newblock {\em IEEE Trans. Biomed. Eng.}, 61(10):2593--2601, 2014.

\bibitem{andreia2016TBE}
Andreia~Vieira Moc¸o∗, Stuijk Sander, and Gerard de Haan.
\newblock Ballistocardiographic artifacts in ppg imaging.
\newblock {\em IEEE Trans. Biomed. Eng.}, 63(9), 2016.

\bibitem{niu2018Synrhythm}
Xuesong Niu, Hu Han, Shiguang Shan, and Xilin Chen.
\newblock Synrhythm: Learning a deep heart rate estimator from general to
  specific.
\newblock In {\em Proc. IAPR ICPR}, pages 3580--3585, 2018.

\bibitem{niu2018VIPL-HR}
Xuesong Niu, Hu Han, Shiguang Shan, and Xilin Chen.
\newblock {VIPL-HR}: A multi-modal database for pulse estimation from
  less-constrained face video.
\newblock In {\em Proc. ACCV}, pages 562--576, 2018.

\bibitem{niu2019rhythmnet}
Xuesong Niu, Shiguang Shan, Hu Han, and Xilin Chen.
\newblock Rhythmnet: End-to-end heart rate estimation from face via
  spatial-temporal representation.
\newblock {\em IEEE Trans. Image Processing}, 2019.

\bibitem{niu2019robust}
Xuesong Niu, Xingyuan Zhao, Hu Han, Abhijit Das, Antitza Dantcheva, Shiguang
  Shan, and Xilin Chen.
\newblock Robust remote heart rate estimation from face utilizing
  spatial-temporal attention.
\newblock {\em Proc. IEEE FG}, 2019.

\bibitem{poh2010non}
Ming-Zher Poh, Daniel~J McDuff, and Rosalind~W Picard.
\newblock Non-contact, automated cardiac pulse measurements using video imaging
  and blind source separation.
\newblock {\em Opt. Express}, 18(10):10762--10774, 2010.

\bibitem{poh2011advancements}
Ming-Zher Poh, Daniel~J McDuff, and Rosalind~W Picard.
\newblock Advancements in noncontact, multiparameter physiological measurements
  using a webcam.
\newblock {\em IEEE Trans. Biomed. Eng.}, 58(1):7--11, 2011.

\bibitem{philip2018remote}
Philipp~V. ROUAST, Marc~T.P. ADAM, Raymond CHIONG, David CORNFORTH, and Ewa
  LUX.
\newblock Remote heart rate measurement using low-cost rgb face video: a
  technical literature review.
\newblock {\em Frontiers of Computer Science (electronic)}, 2018.

\bibitem{sebastian2018cardiovascular}
Alexander Trumpp Daniel~Wedekind Sebastian, Zaunseder and Malberg Hagen.
\newblock Cardiovascular assessment by imaging photoplethysmography - a review.
\newblock In {\em Biomed. Eng. - Biomed. Tech.}, 2018.

\bibitem{trumpp2017skin}
A. Trumpp, S. Rasche, D. Wedekind, M. Schmidt, Waldow, Gaetjen T., and
  Zaunseder F.
\newblock Skin detection and tracking for camera-based photoplethysmography
  using a bayesian classifier and level set segmentation.
\newblock In {\em Bildverarbeitung für die Medizin}, pages 43--48, 2017.

\bibitem{verkruysse2008remote}
Wim Verkruysse, Lars~O Svaasand, and J~Stuart Nelson.
\newblock Remote plethysmographic imaging using ambient light.
\newblock {\em Opt. Express}, 16(26):21434--21445, 2008.

\bibitem{wang2017algorithmic}
Wenjin Wang, Albertus~C den Brinker, Sander Stuijk, and Gerard de Haan.
\newblock Algorithmic principles of remote ppg.
\newblock {\em IEEE Trans. Biomed. Eng.}, 64(7):1479--1491, 2017.

\bibitem{yang2017estimating}
Cheng Yang, Gene Cheung, and Vladimir Stankovic.
\newblock Estimating heart rate and rhythm via 3{D} motion tracking in depth
  video.
\newblock {\em IEEE Trans. Multimedia}, 19(7):1625--1636, 2017.

\bibitem{yu2019remote}
Zitong Yu, Xiaobai Li, and Guoying Zhao.
\newblock Remote photoplethysmograph signal measurement from facial videos
  using spatio-temporal networks.
\newblock {\em Proc. BMVC}, 2019.

\bibitem{yu2019ICCV}
Zitong Yu, Wei Peng, Xiaobai Li, Xiaopeng Hong, and Guoying Zhao.
\newblock Remote heart rate measurement from highly compressed facial videos:
  An end-to-end deep learning solution with video enhancement.
\newblock In {\em Proc. IEEE ICCV}, October 2019.

\end{thebibliography}
}

\end{document}